\documentclass[remotesensing,article,accept,moreauthors,pdftex]{Definitions/mdpi}
\usepackage{float}
\usepackage{subcaption}

\firstpage{1}
\pubvolume{12}
\issuenum{16}
\articlenumber{2636}
\pubyear{2020}
\copyrightyear{2020}
\history{Received: 26 June 2020; Accepted: 07 August 2020; Published: 15 August 2020}
\updates{yes} % If there is an update available, un-comment this line

\pdfoutput=1

\Title{SAR Image Despeckling by Deep Neural Networks:
from a Pre-Trained Model to an End-to-End Training~Strategy}%Attention AE/ME. The following layout issues have not been checked by the English Editing Department and must be carefully verified by the AE/Layout Department: All callout issues, bold usage of callouts, and references to callouts in the text. Correct callout usage in figures. Figure and Table layout issues. Footnote formatting and Glossaries have not been checked. En dash usage for negative values, en dash usage to indicate relationships, en dash usage to indicate bonds (especially in chemistry). The English Editing Department is not responsible for correct italic usage for genes, proteins and technical terminology. This responsibility belongs to the authors. The following are also not checked: spacing between numbers and units of measurement, ratios, en dashes for ranges, date and time formats, punctuation in equation lines, and less than/more than spacing (< >). Finally, capitalization and layout of titles/headings must be properly checked as well as ensuring 'Eq.' and 'Fig.' are properly spelled out, as these are layout issues.

\Author{Emanuele Dalsasso $^{1,*}$\href{https://orcid.org/0000-0001-7170-9015}{\orcidicon}, Xiangli Yang $^{1,2}$\orcidB{}, Lo{\"i}c Denis $^{3}$\orcidC{}, Florence Tupin $^{1}$\orcidD{} and Wen Yang $^{2}$\href{https://orcid.org/0000-0002-3263-8768}{\orcidicon}}

\AuthorNames{Emanuele Dalsasso, Xiangli Yang, Lo{\"i}c Denis, Florence Tupin and Wen Yang}

\address{%
$^{1}$ \quad LTCI, Télécom Paris, Institut Polytechnique de Paris, F-91120 Palaiseau, France; xiangliyang@whu.edu.cn~(X.Y.); florence.tupin@telecom-paris.fr~(F.T.)\\
$^{2}$ \quad Electronic Information School, Wuhan University, Wuhan 430072, China; yangwen@whu.edu.cn\\
$^{3}$ \quad Univ Lyon, UJM-Saint-Etienne, CNRS, Institut d Optique Graduate School, Laboratoire Hubert Curien UMR~5516, F-42023 Saint-Etienne, France; loic.denis@univ-st-etienne.fr}
\corres{Correspondence: emanuele.dalsasso@telecom-paris.fr}

\abstract{Speckle reduction is a longstanding topic in synthetic aperture radar (SAR) images. Many~different schemes have been proposed for the restoration of intensity SAR images. Among~the different possible approaches, methods based on convolutional neural networks (CNNs) have recently shown to reach state-of-the-art performance for SAR image restoration. CNN training requires good training data: many pairs of speckle-free/speckle-corrupted images. This is an issue in SAR applications, given the inherent scarcity of speckle-free images. To handle this problem, this paper analyzes different strategies one can adopt, depending on the speckle removal task one wishes to perform and the availability of multitemporal stacks of SAR data. The first strategy applies a CNN model, trained to remove additive white Gaussian noise from natural images, to a recently proposed SAR speckle removal framework: MuLoG (MUlti-channel LOgarithm with Gaussian denoising). No training on SAR images is performed, the network is readily applied to speckle reduction tasks. The second strategy considers a novel approach to construct a reliable dataset of speckle-free SAR images necessary to train a CNN model. Finally, a hybrid approach is also analyzed: the CNN used to remove additive white Gaussian noise is trained on speckle-free SAR images. The proposed methods are compared to other state-of-the-art speckle removal filters, to evaluate the quality of denoising and to discuss the pros and cons of the different strategies. Along with the paper, we make available the weights of the trained network to allow its usage by other researchers.}

%; MuLoG.}

\newcommand{\proba}{\text{p}}
\newcommand{\cleanimage}{X}
\newcommand{\noisyimage}{Y}
\newcommand{\noise}{N}

\renewcommand{\cleanimage}{x}
\renewcommand{\noisyimage}{y}
\renewcommand{\noise}{n}
\newcommand{\logtransformed}[1]{\widetilde{#1}}
\newcommand{\homomorphic}{homom.-CNN}
\newcommand{\V}[1]{\boldsymbol{#1}}

\begin{document}
%%%%%%%%%%%%%%%%%%%%%%%%%%%%%%%%%%%%%%%%%%

%%%%%%%%%%%%%%%%%%%%%%%%%%%%%%%%%%%%%%%%%%
\section{Introduction}
Synthetic Aperture Radar (SAR) provides high-resolution, day-and-night and weather-independent images.~SAR technology is widely used for remote sensing in Earth observation applications.~With~advanced techniques, like polarimetry, interferometry and differential interferometry, SAR images have numerous applications, ranging from environmental system monitoring, city sustainable development, disaster detection applications up to planetary exploration~\cite{moreira2013tutorial}. SAR is an active system that makes measurements by illuminating a scene and measuring the coherent sum of several backscattered echoes. As~such, the~measured signal suffers from strong fluctuations, that appears in the images as a granular ``salt and pepper'' noise: the speckle phenomenon. The~presence of speckle in an image reduces the ability of a human observer to resolve fine details within the image~\cite{lee1994speckle} and impacts automatic image analysis~tools.

It is well-known that the speckle phenomenon is caused by the presence of many elemental scatterers within a resolution cell, each back-scattering an echo with a different phase shift. The~coherent summation of all these echoes produces strong fluctuations of the resulting intensity from one cell to the next~\cite{goodman1976some}. Speckle analysis and reduction is a longstanding topic in SAR imagery. The~literature on this topic is extensive and goes from single polarization data to fully polarimetric SAR images (see the reviews~\cite{touzi2002review,argenti2013tutorial}). Among~the different possible strategies, we classify some commonly used despeckling algorithms into the following three~categories.

\textit{(i) Selection based methods.} The simplest way to reduce SAR speckle is to average neighboring pixels within a fixed window. This technique is called spatial multilooking. To~reduce the degradation of the spatial resolution when applied to edges, lines or point-link scatterers, both local~\cite{lee1999polarimetric,lopes1993structure,feng2011sar} and non-local  approaches~\cite{deledalle2009iterative,deledalle2011nl,chen2011nonlocal,deledalle2015nl} have been~proposed.

\textit{(ii) Variational methods.} These methods formulate the restoration problem as an optimization problem, to~find the underlying image which best explains the observed speckle-corrupted image. The~objective function to be minimized is typically composed of two terms: a data-fitting term (that~uses a Gamma~\cite{baraldi1995refined}, Rayleigh~\cite{kuruoglu2004modeling} or Fisher--Tippett distribution~\cite{bioucas2010multiplicative} to model the distribution of SAR images) and a regularization term (the total variation TV has been widely used in the literature~\cite{aubert2008variational,darbon2007use,shi2008nonlinear,denis2009sar,bioucas2010multiplicative}).

\textit{(iii) Transform based methods.} Various scientific methods have considered the application of a~wavelet transform for despeckling~\cite{argenti2002speckle, dai2004bayesian, bianchi2008segmentation, li2013bayesian}, but~the estimation of the parameters of the signal and noise statistics is a difficult task. To~apply the state-of-the-art methods designed for additive-noise removal in a more straightforward fashion, a~logarithmic transformation (\textit{a.k.a.} a homomorphic transform) is often applied to convert the multiplicative behavior of speckle into an additive component. Xie~et~al.~\cite{xie2002statistical} have derived the statistical distribution of log-transformed speckle noise. Many~despeckling methods based on a homomorphic transform have been proposed~\cite{parrilli2012nonlocal, achim2003sar, solbo2004homomorphic, xie2002sar, bhuiyan2007spatially}. Special care must be taken because, after~a~log-transform, the~speckle is stationary but not Gaussian and not even centered. At~least a~debiasing step must be included to account for the expectation of log-transformed speckle, as~proposed in~\cite{deledalle2017mulog} and discussed in Section~\ref{sec:pm22}.

Deep learning allows computational models formed of multiple processing layers to learn representations of data that include several levels of abstraction~\cite{lecun2015deep}. In~recent years, these methods have dramatically improved the state-of-the-art in many computer vision fields, such as object detection~\cite{ren2015faster}, face recognition~\cite{sun2014deep}, and~low-level image processing tasks~\cite{chan2015pcanet}. Among~them, convolutional neural networks (CNNs) with very deep architecture~\cite{krizhevsky2012imagenet}, displaying a large capacity and flexibility to represent image characteristics, are well-suited for image restoration. Dong~et~al.~\cite{dong2016image} proposed an end-to-end CNN mapping between the low/high-resolution images for single image super-resolution, which demonstrated state-of-the-art restoration quality, and~achieved fast speed for practical on-line usage. CNN networks have also been applied to the denoising of natural images, but~mostly in the context of additive white Gaussian noise (AWGN). Zhang~et~al. proposed a feed-forward denoising convolutional neural networks (DnCNN) to embrace the progress in very deep architecture. The~advanced regularization and learning methods, including Rectifier Linear Unit (ReLU), batch normalization and residual learning~\cite{he2016deep} are adopted, dramatically improving the denoising performance. There are also some researches ongoing on non-AWGN denoising tasks, such as salt-and-pepper noise~\cite{dong2012wavelet}, multiplicative noise~\cite{wang2018deep}, and~blind inpainting~\cite{xie2012image}. For~SAR image despeckling, CNNs have been first used to learn an implicit model in~\cite{chierchia2017sar}. More and more SAR despeckling methods are now based on deep architectures~\cite{wang2017generative, zhang2018learning, boulch2018learning}.

In this paper, we try to shed light on the advantages and disadvantages of making a significant effort to create training sets and learning SAR-specific CNNs rather than readily applying generic networks pre-trained for AWGN removal on natural images to SAR despeckling. We believe that considering this question for intensity images is enlightening for the more difficult case of multi-channel SAR despeckling that arises in SAR polarimetry, SAR interferometry or SAR~tomography.

In order to discuss this matter, we consider two different SAR despeckling frameworks based on CNNs. The~first one consists of applying a CNN pre-trained on AWGN removal from natural images. Many CNNs have been proposed for AWGN suppression. The~extension to SAR imagery requires to account for the statistics of speckle noise. This can be performed by an iterative scheme recently introduced for speckle reduction: MuLoG algorithm~\cite{deledalle2017mulog}, which is based on the plug-in ADMM strategy~\cite{chan2017plug}. In~the second approach, a~network is trained specifically on SAR images. To~that end, we describe a new procedure to generate a high-quality training set. We consider a network architecture similar to that used in the work of Chierchia~et~al.~\cite{chierchia2017sar} and discuss the influence of the number of layers and of the loss function on the despeckling performance of the CNN, trained and tested on our datasets.
The performance of despeckling methods based on deep neural networks depends not only on the network architecture but also on the training set and on the optimization of the network. Published results are then difficult to reprduce, unless~the network architecture and weights are released. To~facilitate comparisons of future despeckling methods with our work, we~provide an open-source code that includes the network weights for Sentinel-1 image despeckling (see~Section~\ref{sec:cl}).
%Moved by the belief that all the community has a stake in using an existing work (e.g. as a benchmark, to do some testing on their own data, etc.), one of the contributes that comes along with this paper is the release of an open-source code of the trained CNN for Sentinel-1 image despeckling (refer to Section~\ref{sec:cl}).
We have also experimented a hybrid approach, in~the sense that the generated SAR dataset has been used to train a CNN for AWGN removal on images whose content is the same as in the task that it will~perform.
% used in the first method in place of the natural images in order to analyze the impact of training a CNN for AWGN removal on images whose content is the same as in the task that it will perform.

The remainder of the paper is organized as follows. Section~\ref{sec:rw} provides a detailed survey of related image denoising works using CNNs. Section~\ref{sec:pm} first introduces the statistics of SAR data, and~then presents the two different SAR despeckling strategies considered in the paper, plus the hybrid approach. In~Section~\ref{sec:er}, extensive experiments are conducted to evaluate restoration performance. Finally, our concluding remarks are given in Sections~\ref{sec:cl} and~\ref{sec:conclusions}.

%%%%%%%%  RELATED WORK  %%%%%%%%%
\section{Related~Works} \label{sec:rw}

The goal of image denoising is to recover a clean image $\cleanimage$ from a noisy observation $\noisyimage$ which follows a specific image degradation model. In~this paper, we discuss two common degradation models: additive noise ($\noisyimage = \cleanimage + \noise$) and multiplicative noise ($\noisyimage = \cleanimage \times \noise$), where $\noise$ is a random component referred to as ``the noise'' (the terminology ``noise'' to describe fluctuations due to speckle is sometimes considered misleading in the context of SAR imaging given that a pair of images acquired under an interferometric configuration have correlated speckle components, which makes it possible to extract meaningful information from the interferometric phase; we will, however, stick to the terminology common in image processing by referring to the speckle as a noise term throughout the paper since we focus on the restoration of intensity-only images).
%MDPI: Footnote is not permitted in this journal, so we have moved it into the text, please confirm.
%ED: confirmed

\subsection{Additive Gaussian Noise Eeduction by Deep~Learning} \label{sec:rw1}

In the additive model, one usual assumption is that the noise component $\noise$ corresponds to an additive white Gaussian noise (AWGN). To~perform the estimation of $\cleanimage$, a~maximum {\it a posteriori} (MAP) estimation is often considered, where the goal is to maximize the posterior probability:
\begin{equation}\label{eq:map}
\begin{aligned}
\hat{\cleanimage} &= \mathop{\arg\max}_{\cleanimage} \ \proba\!\left(\cleanimage|\noisyimage\right) \\
&= \mathop{\arg\min}_{\cleanimage} \ -\log\, \proba\!\left(\noisyimage|\cleanimage\right) - \log \,\proba\!\left(\cleanimage\right),
\end{aligned}
\end{equation}
with $\proba\!\left(\noisyimage|\cleanimage\right)$ the likelihood defining the noise model, and~$\proba\!\left(\cleanimage\right)$ the prior distribution corresponding to the statistical model of clean images. Under~an additive white Gaussian noise assumption, the~term $-\log\, \proba\!\left(\noisyimage|\cleanimage\right)$ takes the form of a sum of squares: $-\log\, \proba\!\left(\noisyimage|\cleanimage\right)=\frac{1}{2\sigma^2}\|y-x\|_2^2$, with~$\sigma$ the standard deviation of the noise. The~MAP estimator then corresponds to:
\begin{equation}\label{eq:map}
\begin{aligned}
\hat{\cleanimage}
&= \mathop{\arg\min}_{\cleanimage} \ \frac{1}{2\sigma^2}\|y-x\|_2^2 + g(\cleanimage)\\
&=\text{prox}_{\sigma^2g}(y)\,,
\end{aligned}
\end{equation}
where $g(x)=- \log \,\proba\!\left(\cleanimage\right)$ and the notation $\text{prox}_{g}$ defines the proximal operator associated to function $g$ \cite{combettes2011proximal,parikh2014proximal}. Classical prior models $g$ include the total variation, sparse analysis and sparse synthesis priors~\cite{elad2007analysis}. While earlier models were handcrafted ($\ell_1$ norm of the wavelet coefficients for a specifically chosen transform, total variation), more recent models are learned from sets of natural images (field of experts~\cite{roth2005fields}, higher-order Markov random fields~\cite{chen2014insights}).

Beyond MAP estimators, several patch-based methods were designed to estimate the clean image $\cleanimage$, such as BM3D~\cite{dabov2007image}, LSSC~\cite{mairal2009non} or WNNM~\cite{gu2014weighted}. These methods exploit the self-similarity observed in most natural images (repetition of similar structures/textures at the scale of a patch of size typically $8\times 8$ within extended neighborhoods).

In order to learn richer patterns of natural images, deep neural networks have been considered for denoising purposes. After~the training step, the~estimation step is very fast, especially on graphical processing units (GPUs).
The first network architectures designed for denoising were trained to learn a mapping from noisy images $\noisyimage$ to clean ones $\cleanimage$ by making use of a CNN with feature maps~\cite{jain2009natural}, a~multi-layer perceptron (MLP) \cite{burger2012image}, a~stacked denoising autoencoder (SDA) network~\cite{vincent2010stacked}, a~stacked sparse denoising auto-encoder architecture~\cite{xie2012image}, among~all.
% Other architectures of network include RED-Net~\cite{mao2016image}, DenoiseNet~\cite{remez2017deep}, and Memnet~\cite{tai2017memnet}.

More recent networks are trained to estimate the noise component $\noise$, i.e.,~to output the residual $\noisyimage-\cleanimage$. This strategy, called ``residual learning'' \cite{he2016deep}, has shown to be more efficient when the neural network contains many layers (i.e., for~deep networks).
In~\cite{zhang2017beyond}, this residual learning formulation for model learning is adopted and 17 layers are used with the $\ell_2$ loss function. Wang~et~al.~\cite{wang2017dilated} replaced the convolution layer by dilated convolution. In~\cite{wang2017elu,isogawa2018deep}, an~exponential linear unit or soft shrinkage function are used as activation function (i.e., the~non-linear step that follows the convolution), instead of the Rectified Linear Unit (ReLU). Liu~et~al.~\cite{liu2017wide} discuss the relation between the width and the depth of the network. They introduce wide inference networks with only 5 layers. In~\cite{bae2017beyond}, a~wavelet transform is introduced into deep residual~learning.

All these variations on a common structure are motivated by finding a trade-off between the network expressivity (in particular, by~increasing the size of the receptive field) and the generalization potential (i.e., fighting against the over-fitting phenomenon that arises when the number of network parameters increases).

Most networks generate an estimate $\hat \cleanimage$ of the clean image from a noisy image $\noisyimage$ by simple traversal of the feedforward network (and possibly subtracting the residuals to the noisy image, in~case of residual learning): the proximal operator $\text{prox}_{g}$ is learned instead of the prior distribution $\proba(\cleanimage)$, no~explicit minimization is performed when estimating a denoised image. The~plug-and-play ADMM strategy~\cite{chan2017plug} provides a means to apply implicit modeling of the prior distribution $\proba(\cleanimage)$ encoded within the network in the form of the proximal operator $\text{prox}_{g}$ in order to address more general image restoration problems than the mere AWGN~removal.

%\IEEEpubidadjcol

\subsection{Speckle Reduction by Deep~Learning} \label{sec:rw2}
In the past years, most of the SAR image denoising approaches are based on detailed statistical models of signal and speckle. To~avoid the problem of modeling the statistical distribution of speckle-free SAR images, several authors recently resorted to machine learning approaches implemented through~CNNs.

The first paper that investigates the problem of SAR image despeckling through CNNs is~\cite{chierchia2017sar}. Following the paradigm proposed in~\cite{zhang2017beyond}, the~SAR-CNN implemented by Chierchia~et~al. is trained in a residual fashion, where a homomorphic approach~\cite{deledalle2017mulog} is used in order to stabilize the variance of speckle noise. The~network comprises 17 convolutional layers with Batch Normalization~\cite{ioffe2015batch} and Rectifier Linear Units (ReLU) \cite{krizhevsky2012imagenet} activation function. Logarithm and hyperbolic cosine are combined in a smoothed $\ell_1$ loss function. The~Image Despeckling Convolutional Neural Network (ID-CNN) proposed by Wang~et~al.~\cite{wang2017sar} comprises only 8 layers and is applied directly on the input image without a log-transform. The~novelty lies in the formulation of the loss function as a combination of an $\ell_2$ loss and of Total Variation (TV), preventing the apparition of artifacts while preserving important details such as edges. In~\cite{zhang2018learning}, the~proposed SAR-DRN makes use of dilated convolutions and skip connections to increase the receptive field without increasing the complexity of the network and maintaining the advantages of 3 $\times$ 3 filters. Wang~et~al.~\cite{wang2017generative}  tackle the image despeckling problem by resorting to generative adversarial networks. In~the proposed ID-GAN, the~Generator is trained to directly estimate the clean image from a noisy observation, while the discriminator serves to distinguish the de-speckled image synthesized by the generator from the corresponding ground truth image. To~exploit the abundance of stacks of multitemporal data and avoid the problem of creating a clean reference, in~\cite{boulch2018learning} a CNN is used as an auto-encoder through a formulation that does not use an explicit expression of the noise model, thus allowing~generalization.

%%%%%%%%  PROPOSED METHOD  %%%%%%%%%
\section{SAR Despeckling Using~CNNs}
%convolutional neural network}
\label{sec:pm}
\vspace{-6pt}

\subsection{Statistics of SAR~Images} \label{sec:pm1}

After SAR focusing, the~SAR image is formed by the collection of the complex amplitudes back-scattered by each resolution cell. The~squared modulus of this complex amplitude (the~\emph{intensity} image) is informative of the total reflectivity of the scatterers in each resolution cell. Because~of the interference between echoes produced by elementary scatterers, the~intensity fluctuates (speckle~phenomenon). These fluctuations depend on the 3-D spatial configuration of the scatterers with respect to the SAR system and on the nature of the scatterers. They are generally modeled by Goodman's stochastic model (fully developed speckle): the measured intensity $\noisyimage$ is related to the reflectivity $\cleanimage$ by the multiplicative model $\noisyimage = \cleanimage \times \noise$ where the noise component $\noise\in\mathbb{R}^+$ is a random variable that follows a gamma distribution~\cite{goodman1976some}:
\begin{equation}\label{eq:pdf_int}
\proba(\noise) =\frac{L^L}{\Gamma(L)} \noise^{L-1}\exp\left(-L\noise\right)\,,
\end{equation}
with $L\geq 1$ representing the number of looks, and~$\Gamma(\cdot)$ the gamma function. It follows from $\mathbb{E} [\noise] = 1$ and $\text{Var}[\noise] = 1/L$ that $\mathbb{E} [\noisyimage] = \cleanimage$: averaging the intensity leads to an unbiased estimator of the reflectivity in stationary areas, and~$\text{Var}[\noisyimage] = \cleanimage^2/L$: the noise is signal-dependent in the sense that the variance of the measured intensity increases like the square of the~reflectivity.

Considering the log of the intensity $\logtransformed{\noisyimage}$ instead of the intensity $\noisyimage$ transforms the noise into an~additive component: $\logtransformed{\noisyimage}=\logtransformed{\cleanimage}+\logtransformed{\noise}$ where $\logtransformed{\noise}\in\mathbb R$ follows a Fisher-Tippett distribution~\cite{xie2002statistical}:
\begin{equation}\label{eq:pdf_logint}
\proba(\logtransformed{\noise}) = \frac{L^L}{\Gamma(L)}e^{L\logtransformed{\noise}}\exp(-Le^{\logtransformed{\noise}}).
\end{equation}
The mean $\mathbb{E} [\logtransformed{\noise}] = \psi(L)-\log(L)$ ($\psi$ is digamma function) is non-zero, which indicates that averaging log-transformed intensities $\logtransformed{\noisyimage}$ leads to a biased estimate of the log-reflectivity $\logtransformed{\cleanimage}$. The~variance $\text{Var}[\logtransformed{\noise}]~=~\psi(1,L)$ (where $\psi(1, L)$ is the polygamma function of order $L$ \cite{abramowitz1965handbook}) does not depend on the reflectivity: it is stationary over the~image.

\subsection{Despeckling Using Pre-Trained CNN~Models} \label{sec:pm2}

As discussed in Section~\ref{sec:rw}, several approaches have been
recently proposed in the literature to apply CNNs to AWGN
suppression. The~residual learning method DnCNN introduced in
~\cite{zhang2017beyond} is a~reference method for which models pre-trained
on natural images at various signal-to-noise ratios are
available ({\url{https://github.com/cszn/DnCNN}}).
%remote sensing do not  have footnote.
We consider
two different ways to apply a pre-trained CNN to speckle noise
reduction: a homomorphic filter that processes log-transformed
intensities with DnCNN and the embedding of DnCNN within the iterative
scheme MuLoG~\cite{deledalle2017mulog}. Note that our approach is
general and other pre-trained CNNs than the DnCNN could readily be
applied.

\medskip

\subsubsection{Architecture of the~CNN} \label{sec:pm21}
\begin{figure}[H]
\centerline{
\includegraphics[width=15.4 cm]{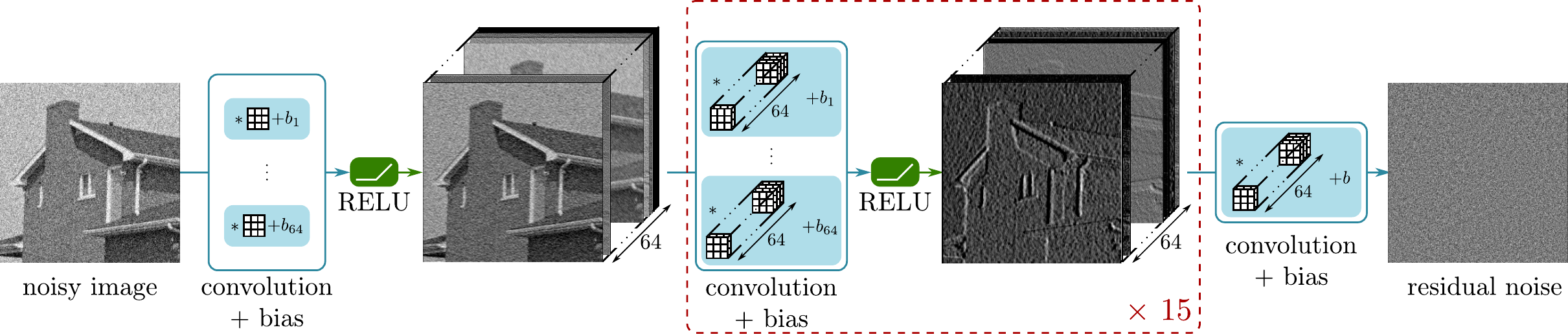}
}
\caption{The architecture of DnCNN proposed in~\cite{zhang2017beyond}.}
\label{fig:dncnn}
\end{figure}

Figure~\ref{fig:dncnn} and Table~\ref{tab:DnCNN} illustrate the architecture of the
network DnCNN proposed by Zhang~et~al.~\cite{zhang2017beyond}.
The network is a modified VGG network and is made of 17 fully
convolutional layers with no pooling. There are three types of layers:
(i) \textit{Conv+ReLU}: for the first layer, 64 filters of size
3~$\times$~3 are used to generate 64 feature maps, and~rectified linear
units are then utilized for nonlinearity; (ii) \textit{Conv+BN+ReLU}:
for layers 2$\sim$16, 64 filters of size 3~$\times$~3~$\times$~64 are
used, and~batch normalization is added between convolution and ReLU;
(iii) \textit{Conv}: for the last layer, a~filter of size
$3\times3\times64$ is used to reconstruct the output. The~loss
function that is minimized during the training step is the $\ell_2$
loss (i.e.,~the~sum of squared errors, averaged over the whole
training set). To~train the DnCNN, 400 natural images of size 180$\times$180 pixels
with gray levels in the range $[0,1]$ were used for
training. Patches of size $40\times 40$ pixels were extracted at
random locations from these images. Different networks were trained
for simulated additive white Gaussian noise levels equal to $\sigma =
\tfrac{10}{255},\tfrac{15}{255},...,\tfrac{75}{255}$ (14 different
networks each corresponding to a given noise standard deviation).

\unskip

\begin{table}[H]
\centering
\caption{Configuration of a DnCNN with D~layers.}
\begin{tabular}{ccc}
\toprule
\textbf{Name} & $\textbf{\emph{N}}_{\textbf{\emph{out}}}$ & \textbf{Configuration}  \\
\midrule
Layer 1 & 64 & $3 \times 3$ CONV, ReLU \\
Layer 2 to (D-1) & 64 & $3 \times 3$ CONV, Batch Norm., ReLU \\
Layer D & 1 & $3 \times 3$ CONV \\
\bottomrule
\end{tabular} \label{tab:DnCNN}
\end{table}
\unskip

\subsubsection{Homomorphic Filtering with a Pre-Trained~CNN} \label{sec:pm22}

\begin{figure}[H]
\centering
{\includegraphics[width=.8\textwidth]{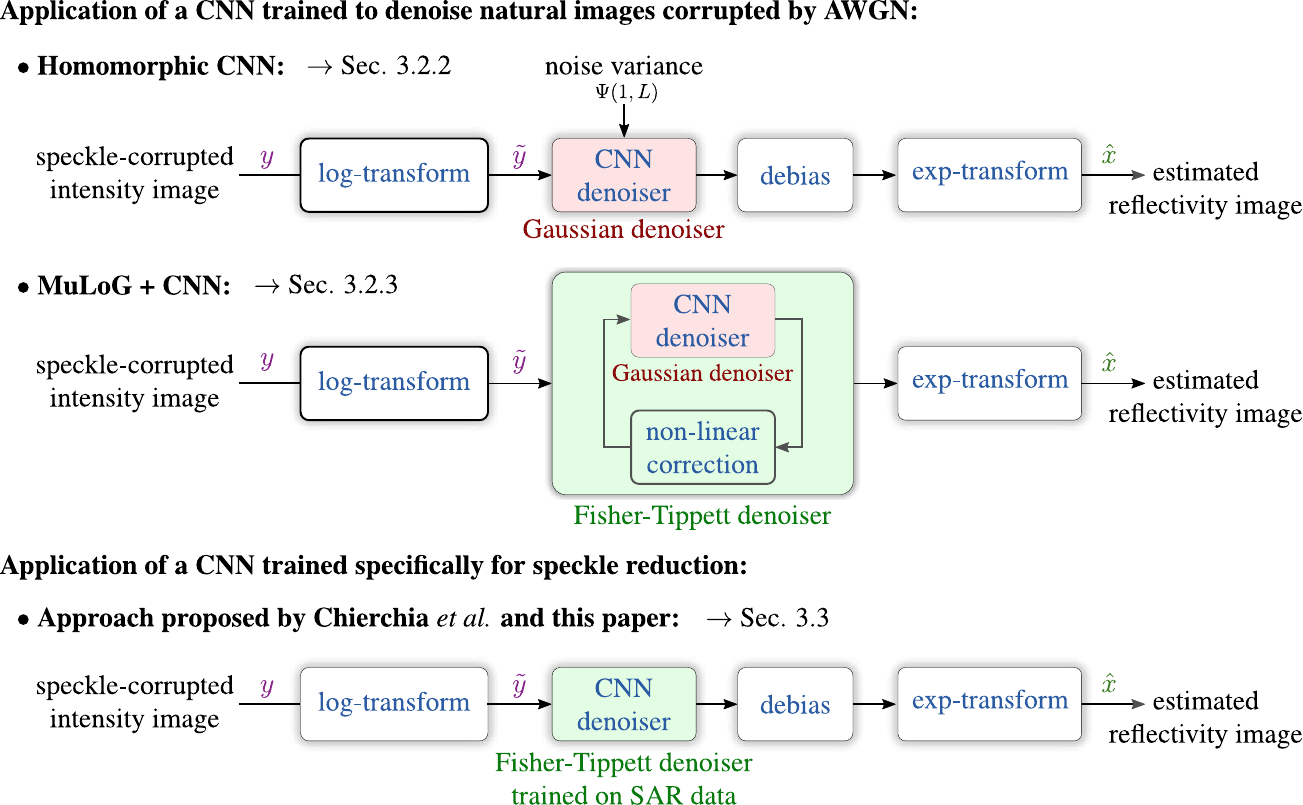}}
\caption{Illustration of the three speckle reduction approaches described
in this paper: the first two apply a CNN trained to remove AWGN from
natural images, the~last approach consists of training a~CNN
specifically to the suppression of speckle in (log-transformed) SAR images.
}
\label{fig:hcnn}
\end{figure}

The simplest approach to applying a pre-trained CNN acting as a Gaussian
denoiser is the homomorphic filtering depicted at the top of
Figure~\ref{fig:hcnn}, and~hereafter denoted \homomorphic.
This~approach consists of approximating the noise term
$\logtransformed{\noise}$ in log-transformed data as an additive white
Gaussian noise with non-zero mean. As~recalled in
Section~\ref{sec:pm1}, log-transformed speckle is not Gaussian but follows a
Fisher-Tippett distribution under Goodman's fully developed speckle
model. Hence, the~homomorphic approach is built on a rather coarse
statistical approximation. Under~this approximation, the~log-transformed SAR image can be restored by first applying the
pre-trained CNN, then correcting for the bias $\psi(L)-\log L$ due to
the non-centered noise~component.

In order to successfully apply a pre-trained CNN model, it is crucial
to properly normalize the data so that the range of input data matches
the range of data used during the training step (neural~networks are
highly non-linear). This paper maps the range $[q_m,\,q_M]$ by an affine
transform to the $[0,1]$ range, with~$q_m$ and $q_M$ corresponding to
the $0.3\%$ and $99.7\%$ quantiles of the log-transformed
intensities. After~this normalization, the~standard deviation of the
log-transformed noise is \mbox{$\sigma~=~\sqrt{\psi(1,L)}/(q_M-q_m)$}. This
value can be used to select the network trained for the closest noise
standard deviation $\sigma_{\text{train}}$ that is less or equal to
$\sigma$. The~normalized image is then multiplied by
$\sigma_{\text{train}}/\sigma$ so that the noise standard deviation
exactly matches that of the images in the training set of the~network.

\subsubsection{Iterative Filtering with MuLoG and a Pre-Trained~Model} \label{sec:pm23}
\begin{figure}[H]
\centering
\includegraphics{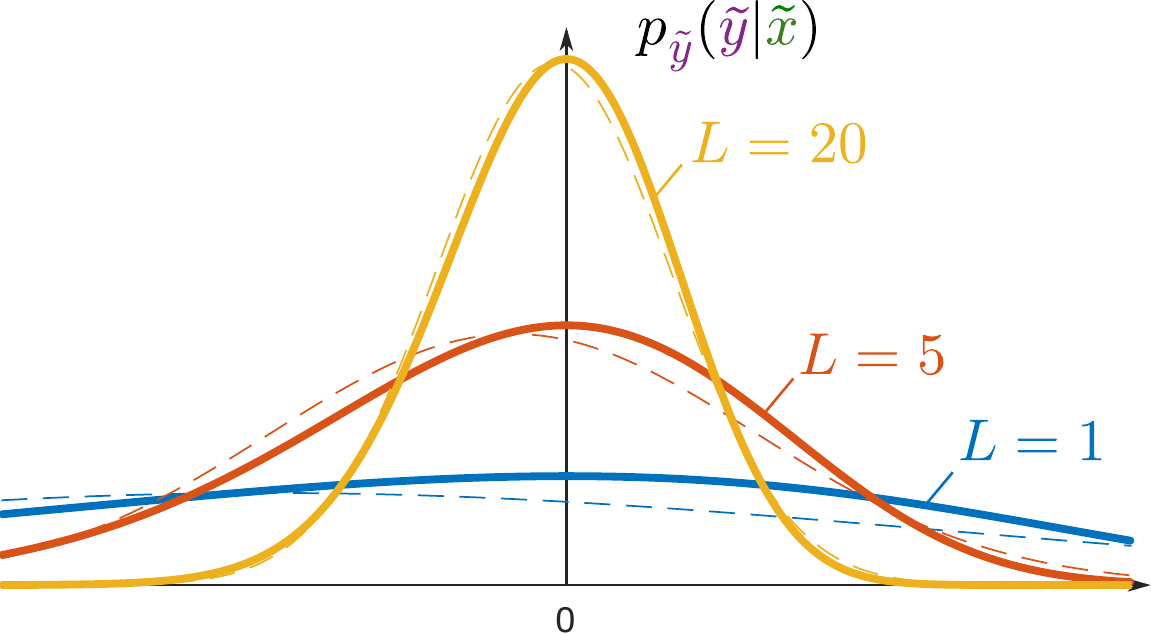}
\caption{Comparison of the Fisher-Tippett distribution (with different number of looks L, represented by the continuous line) and the Gaussian distribution (dashed line) with the same mean and~variance.}
\label{fig:logstats}
\end{figure}

The MuLoG framework~\cite{deledalle2017mulog} accounts for the
Fisher-Tippett distribution of log-transformed speckle (see Figure~\ref{fig:logstats}) with an
iterative scheme that alternates the application of a Gaussian
denoiser (namely, a~proximal operator) and of a non-linear correction.
Figure~\ref{fig:hcnn}, second row, illustrates that, by~embedding a
CNN trained as a Gaussian denoiser within an iterative scheme, a~Fisher-Tippett denoiser is obtained.
Throughout the iterations, the~parameter of the
Gaussian denoiser evolves, which~requires to apply the network
selection and image normalization strategy described in the previous~paragraph.

\unskip

\subsection{Despeckling with a CNN Specifically Trained on SAR~Images} \label{sec:pm3}
The architecture that we consider for our CNN trained on SAR images
(SAR-CNN) is
based on the work of Chierchia~et~al.
~\cite{chierchia2017sar}, who in turn was inspired by the DnCNN by
Zhang~et~al. \cite{zhang2017beyond} that we described in
Section~\ref{sec:pm21}. In~this section, we describe in details a procedure
for producing high-quality ground-truth images for the training step
of the~network.

%an \textit{ad-hoc} ground truth-generation procedure, consisting in temporal multilooking followed by a MuLoG+BM3D denoising step. The image obtained is the target of the deployed CNN-based approach. Synthetic SAR images in amplitude format, generated by multiplying with 1-look speckle noise the clean references, feed the network, which has the goal of reproducing the noise affecting the input image. This approach is called residual learning. The estimated noise is then removed from the image to obtain the estimated clean image.

\medskip
\subsubsection{Training-Set~Generation} \label{sec:pm31}
Deep learning models need a lot of data to generalize well. This is an
issue in deep learning-based SAR image despeckling techniques, due to
the lack of truly speckle-free SAR images. The~reference image has to be
therefore created through an \textit{ad-hoc} procedure, in~order to produce images taking into account the content of SAR data (strong backscattering scatterers, real radiometric
contrasts of SAR data). This is done by exploiting series of SAR~images.

Speckle noise can be strongly reduced by
multi-temporal multilooking (i.e., averaging the intensity of images
acquired at different dates, assuming that no changes occurred). Multi temporal stacks have therefore been considered in this study. Due to the coherence of some regions, some speckle
fluctuations are remaining after this temporal multi-looking
procedure. These images are further improved by applying a MuLoG+BM3D
denoising step~\cite{deledalle2017mulog} with an equivalent
number of looks estimated from selected homogeneous regions. The~images obtained are then considered speckle-free
and serve as a ground truth. Synthetic speckle noise is simulated
based on the statistical models described in Section~\ref{sec:pm1} in
order to produce the noisy/clean image pairs necessary for the
training of the network. Although~Goodman's fully developed model is not verified everywhere in the images (the Rician distribution~\cite{Goodman,Elto-05,nicolas2019rice} could instead be used for strong scatterers) and assumes spatially uncorrelated speckle, it has been the funding model of most SAR despeckling methods developed these last four decades~\cite{aubert2008variational}. Thus, we find it  relevant to simulate 1-look speckle noise based on Goodman's model. In~Section~\ref{sec:er3}, its limitations are~discussed.

\begin{figure}[H]
\centering
%	\captionsetup{justification=centering}
\includegraphics[width=14.1 cm]{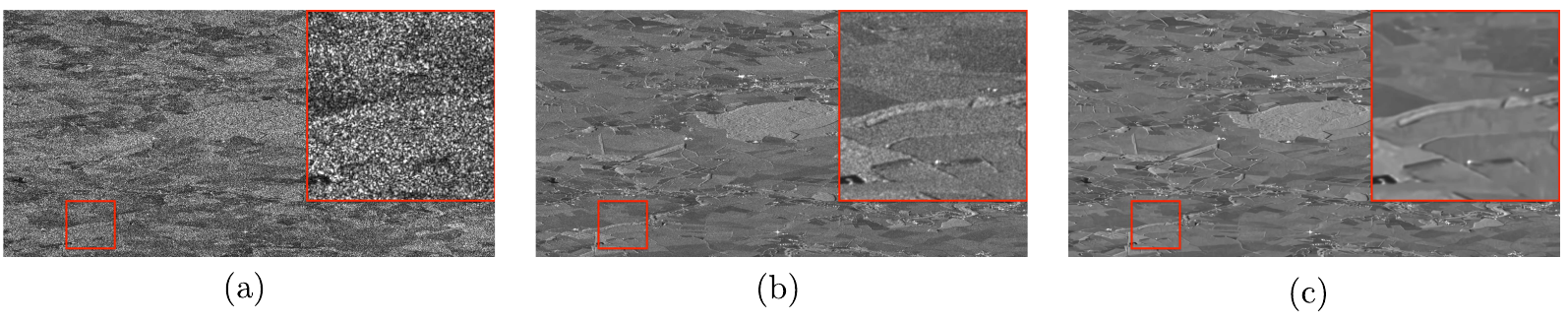}
\caption{(\textbf{a}) A 1-look SAR image acquired by Sentinel-1 \copyright ESA, (\textbf{b}) result of multitemporal averaging of 45 dates, (\textbf{c}) virtually speckle-free image obtained by denoising image (\textbf{b}) with MuLoG+BM3D.}
\label{fig:cleaning_process}
\end{figure}

A ground truth image is depicted
in Figure~\ref{fig:cleaning_process}, where we show the progress from
the 1-look SAR image to the clean reference used in our training step. The~temporal average of large temporal stacks of finely registered SAR images leads to images with limited speckle fluctuations and, after~denoising, these images retain the characteristics of SAR images (bright points, sharp edges, textures) with almost no residual fluctuations due to speckle. Since the denoising operation is applied on an image already temporally multilooked, only small fluctuations have to be suppressed by the denoising step (i.e., this~denoising step helps but is not crucial). This method thus proposes a realistic way to create speckle-free SAR images. A~description of the training set is then given in Table~\ref{table:training-set_description}.
%The noisy instances are then obtained by adding simulated 1-look speckle noise. Simulated noisy SAR images are thus obtained, along with their corresponding ground truth.

\begin{table}[H]
\centering
\caption{Description of the training-set of the proposed SAR-CNN. For~each image, the~number of dates composing the multi-temporal stack and the number of patches extracted are~given.}
\begin{tabular}{l c c c c}
\toprule
\textbf{Images}      & \textbf{Number of Dates}           & \textbf{Number of Patches} \\
\midrule
Marais 1    & 45 & 40194\\
Limagne     & 53 & 40194 \\
Saclay      & 69 & 7227\\
Lely        & 25 & 14850\\
Rambouillet & 69 & 39168\\
Risoul      & 72 & 9648 \\
Marais 2    & 45 & 40194 \\
\bottomrule
\end{tabular}
%The average PSNR and its variance are separated by semicolon.}
\label{table:training-set_description}
\end{table}
%MDPI: Please confirm if it is necessary to add commas for numbers of more than four-digit/five.
%ED: it is fine without commas

\vspace{-9pt}

\subsubsection{Network Architecture and the Effect of the Loss~Function} \label{sec:pm32}
The network is easier to train on log-transformed data since the noise
is then additive and white (and coarsely Gaussian
distributed, as~it ca be seen from Figure~\ref{fig:logstats}). Compared to the suppression of AWGN, the~networks need
to learn how to separate log-transformed SAR reflectivities from log-transformed speckle, distributed
according to the Fisher-Tippett distribution given in
Equation~(\ref{eq:pdf_logint}). In~the follwing, a~discussion about minor changes to the network architecture is carried out. It is worth to point out that, given the impossibility to reproduce the work proposed in~\cite{chierchia2017sar} (the weights are not available and the datasets are not the same), we intend by no means to compare our results to those of Chierchia~et~al. Instead, we have always followed our training strategy and drawn conclusions based on visual inspection of our testing~set.

%As said in Section~\ref{sec:pm1}, the log transform has the property of stabilize the variance, i.e.,~the fluctuations are made signal independent~\cite{deledalle2017mulog}. After the log is applied to the amplitude images, a normalization step is performed on the whole dataset (training+test images). The normalization between 0 and 1 of the data takes into account the statistics of the entire dataset; namely, the normalization factor is the same for all the images. Denoting with $M$ and $m$ the maximum and minimum value, respectively, and recalling that $X_{HA}=\log X_A$, then
%\begin{equation}
%M=\max_{i}{X_{HA}^i}, \quad m = \min_{i}{X_{HA}^i}, \quad i=1,\dots,N
%\end{equation}
%where $N$ is the number of training images. Normalization of image $X_{HA}$ in amplitude format is defined as:
%\begin{equation}
%\bar{X}_{HA}^i=\bigg(\frac{X_{HA}^i-m }{M-m}\bigg).
%\end{equation}

%The normalized log of the amplitude of the noisy observation, denoted with $\bar{Y}_{HA}^i$ is the input of our neural network. The network is a modified version of the VGG network~\cite{simonyan2014very}, made suitable for image denoising task. Network depth has been set according to the size of the patches used in state-of-the-art denoising algorithms. The framework we rely on in our deep learning algorithm is the residual learning formulation. In our model, batch normalization has been incorporated for fast and stable training and improved performance.

\begin{figure}[H]
\centering
\captionsetup{justification=centering}
\includegraphics[width=15.5 cm]{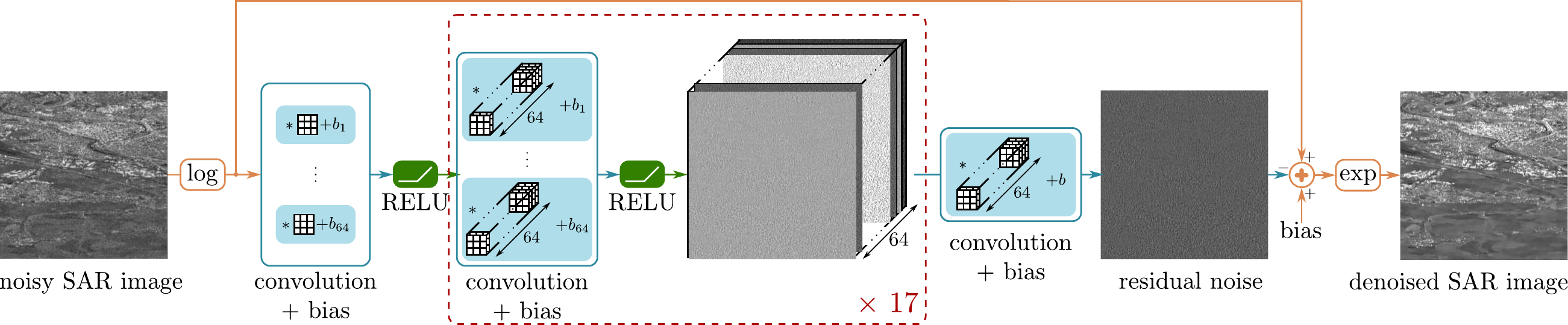}
\caption{\label{fig:sarcnn}The proposed SAR-CNN for Sentinel-1
image despeckling.}
\end{figure}

% NETWORK DEPTH
In this study, it has been experimentally found that increasing the depth of the network
compared to the depth used by Zhang~et~al. and Chierchia~et~al. was improving the performance on the testing set. We used 19
layers, each layer involving spatial convolutions with $3\times 3$
kernels (see Figure~\ref{fig:sarcnn}). The~receptive field of our
network then corresponds to a patch of size $39 \times 39$.\unskip

%
%Following the principle in~\cite{simonyan2014very}, filter size has been set to $3 \times 3$, removing all pooling layers. The receptive field of such network with depth $d$ should be $(2d+1)\times(2d+1)$.
%
%The depth of our SAR-CNN for image despeckling is set to 19, thus resulting in a receptive field of $39 \times 39$ with patches of size $40 \times 40$. Thus, the context information captured by the network to generate the despeckled output is basically all the information contained in the input patch.
%
% LOSS FUNCTION

%The network has a depth of 19 layers. It comprises three types of layers, already described in Section~\ref{sec:pm2}: one \textit{Conv+ReLU} layer is followed by 17 \textit{Conv+BN+ReLU} layers and one \textit{Conv} layer. The network then outputs the residual noise, which is subtracted from the input and converted back to the amplitude domain by inverse normalization and the exp operation.

%Let us recall some terminology. Given a pair of clean-noisy training patches in amplitude format $(X_{A}^i,Y_{A}^i)$, it holds that
%\begin{equation}
%Y_{A}^i = X_{A}^i \times S_{A}^i,
%\end{equation}
%thus
%\begin{equation}
%\frac{Y_A^i}{X_A^i}=S_A^i.
%\end{equation}
%In the log domain, we have that
%\begin{equation}
%Y_{HA}^i - X_{HA}^i = S_{HA}^i.
%\end{equation}
%When normalization is applied, it follows that
%\begin{equation}
%\begin{aligned}
%\bar{Y}_{HA}^i - \bar{X}_{HA}^i &= \bigg(\frac{X_{HA}^i-m }{M-m}\bigg)-\bigg(\frac{Y_{HA}^i-m }{M-m}\bigg) \\
%&=\frac{S_{HA}^i}{M-m} \\
%&=\bar{S}_{HA}^i
%\end{aligned}
%\end{equation}
%
%This approach leads to the following loss function:
To train the network, several loss functions have been considered. Experiments suggest the $\ell_1$ loss function to be preferable to the smoothed $\ell_1$ loss function of Chierchia~et~al. and to $\ell_2$ loss
function. This matches other studies that have shown a reduction of
artifacts and an improvement of the convergence when using the
$\ell_1$ loss~\cite{zhao2017loss}.
We, therefore, used the following loss function:
\begin{equation}\label{eq:l1}
%L^{l_1}\big(\Theta\big) =
\sum_{i=1}^{N}\left\|f_{\text{CNN}}(\logtransformed{\V{\noisyimage}}_i)-\logtransformed{\V{\cleanimage}}_i+(\psi(L)-\log(L))\cdot\V{1}\right\|_1
\end{equation}
where the sum is carried over all $N$ images from (a batch sampled from) the training set,
$f_{\text{CNN}}(\cdot)$ represents the action of the CNN on some
log-transformed input data,
the boldface is used to denote images (the $i$-th noisy image
$\V{\noisyimage}_i$ and the corresponding speckle-free reference
$\V{\cleanimage}_i$). The~term \mbox{$(\psi(L)-\log(L))\cdot\V{1}$} is a
constant image that corresponds to the bias correction. Its role is to
center Fisher-Tippett distribution. By~making this term explicit, it
is easier to perform transfer learning, i.e.,~to~re-train a network to
a different number of looks $L$ by warm-starting the optimization from
the values obtained for the previous number of~looks.

%[LOIC: TODO: check that this is true...]

%$\bar{c}$ is the normalized nonzero mean of log speckle~\cite{xie2002statistical}. From Equation~(\ref{eq:stats_logamplitude}):
%\begin{equation}
%\bar{c}=\frac{\frac{1}{2}(\psi(L)-\text{ln}(L))}{M-m}.
%\end{equation}
%The loss function is the average over the number $N$ of examples comprised in one batch. In our case, $N=128$ patches.

%Loss function $l_2$ is responsible of unwanted artifacts in the output image, when used to train a neural network for image restoration purposes. The reason why $l_1$ loss~\cite{zhao2017loss} has been adopted over $l_2$ is the attempt of reducing the artifacts introduce by the $l_2$ loss function. $l_1$ does not over-penalize large errors, since it weighs errors differently than $l_2$ and has different convergence properties. It has been demonstrated in~\cite{zhao2017loss} that it is more likely for $l_2$ to get stuck to a local minimum than for $l_1$ and that their network trained with $l_1$ outperforms the same network trained with $l_2$.

\medskip
\subsubsection{The Training of the~Network} The training set is formed by 7
Sentinel-1 speckle-free images produced by filtering 7 different
multi-temporal stacks of size between $1024\times 1536$ and $1024\times 8192$ pixels, as~described in Section~\ref{sec:pm31}.
The images are selected so that to cover urban areas, forests, a~coast with some water
surfaces, fields and mountainous areas. Patches of size $40\times 40$ are
extracted from these images, with~a stride of 10 pixels between
patches. Mini-batches of 128 patches are used. In~order to improve the
network generalization capability, standard data augmentation
techniques are used: vertical, horizontal flipping and $\pm 90^{\circ}$ and
$180^{\circ}$ rotations are applied on the patches. 11968 batches of 128
patches are processed during an epoch. A~total of 50 epochs were used
with ADAM stochastic gradient optimization method, with~an initial
learning rate of 0.001. The~convergence of the learning and prevention for
over-fitting were checked by monitoring the decrease of the loss
function throughout the epochs as well as the performance over the
test set. The~deep learning framework used for the implementation is tensorflow 1.1.12. Training is carried out with an Intel Xeon CPUat 3.40 GHz and an Nvidia K80 GPU and took approximately 7~h.

\subsection{Hybrid Approach: MuLoG + Trained~CNN} \label{sec:pm4}
A hybrid approach is also considered. In~this method, the~dataset that we have constructed is used to retrain the CNN described in Section~\ref{sec:pm21} to remove Gaussian noise from SAR images. Account for the Fisher-Tippett distribution is also made possible by embedding this network within the MuLoG framework. By~doing so, we aim at investigating the influence of the content of the training images on the restoration~performances.

\section{Experimental~Results} \label{sec:er}

The following paragraphs provide a comparison of the
different strategies for CNN-based despeckling both on images with
simulated speckle and on single-look Sentinel-1 images. First,~the~impact of the loss function and of the number of layers
on the performance of SAR-CNN is illustrated. Network architecture is described in Sections~\ref{sec:pm21} and \ref{sec:pm32} (and more in details in the original article~\cite{zhang2017beyond}). A~graphic illustration is also provided in Figure~\ref{fig:sarcnn}. The~network has been trained in a supervised way using a dataset created as explained in Section~\ref{sec:pm31}. The~three proposed algorithms are summarized in Table~\ref{tab:description_algorithms}.

\begin{table}[H]
\centering
\caption{Description of the proposed~algorithms.}
\begin{tabular}{cccc}
\toprule
\multirow{2}{*}{\textbf{Algorithm}} &\multirow{2}{*}{{\textbf{MuLoG+CNN}}}&{\textbf{MuLoG+CNN}} & \multirow{2}{*}{{\textbf{SAR-CNN}}}  \\
&&\textbf{(Pretrained on SAR)}&\\
\midrule
Input & Natural images& SAR dataset& SAR dataset\\
Noise type &Gaussian & Gaussian & Speckle\\
Architecture & DnCNN, $D=17$&DnCNN, $D=17$
&DnCNN, $D=19$
\\
Loss function & $\ell_2$& $\ell_2$& $\ell_1$ \\
\bottomrule
\end{tabular}

\label{tab:description_algorithms}
\end{table}
\unskip

\subsection{Influence of the Loss Function and of the Network~Depth}
\begin{figure}[H]
\includegraphics[width=15.5 cm]{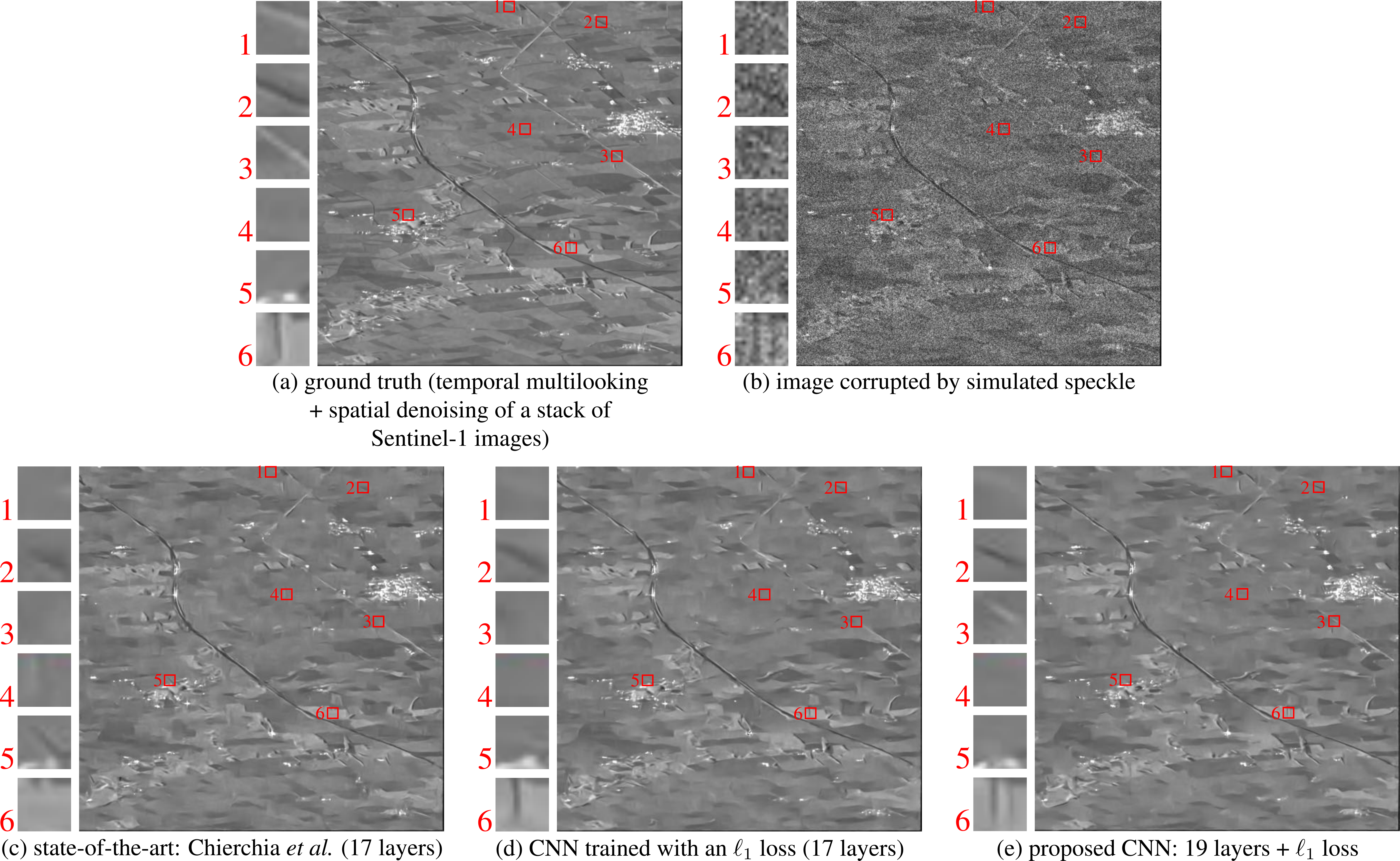}
\caption{Illustration of the influence of the loss function and of the number of layers of the CNN on the despeckling performance: (\textbf{a}) the ground truth image obtained by temporal+spatial filtering of a stack of 45 Sentinel-1 images; (\textbf{b}) the same image corrupted by a synthetic speckle; (\textbf{c}) restoration obtained by training the CNN architecture proposed by Chierchia~et~al. on our training set; (\textbf{d}) restoration obtained when using an $\ell_1$ loss instead of the loss used by Chierchia~et~al.; (\textbf{e}) restoration obtained with an $\ell_1$ loss and two additional~layers.}
\label{fig:comparison0}
\end{figure}

%\subsection{Comparison with previous CNN architectures}
\label{sec:er1}
Figure~\ref{fig:comparison0} shows the impact of the loss
function ($\ell_1$ versus the smoothed $\ell_1$ used by \mbox{Chierchia~et~al.
~\cite{chierchia2017sar}}) and of the network depth in terms
of despeckling artifacts on an image with simulated speckle. The~image corresponds to one of the speckle-free images of our testing set. We~magnify 6 regions in order to illustrate cases where the initial CNN architecture  fails to recover some structures (regions 1, 2, 3 and 6) that are present in the ground truth and at least partially recovered applying the proposed modifications to the CNN, and~cases were some spurious structures appear (regions 4 and 5) with the first CNN architecture employed, but~are not created with the modifications of the CNN depth and loss function that we~considered.

%Since the network coefficients of Chierchia~et~al. are not available, we can only report the results obtained when training an architecture indentical to that of Chierchia~et~al.~\cite{chierchia2017sar}, trained on the high-quality database of SAR images we have created. By changing the depth and loss function, we show how the performance can be improved. All the three networks have been trained on our dataset, using 50 epochs.

All the three architectures are trained for 50 epochs on the high-quality database of SAR images we have created, and~conclusions are drawn after qualitative evaluation of our testing images. Indeed,~as~already claimed, a~comparison with the work of Chierchia~et~al. is not~applicatble.

\unskip

%By visually inspecting the results, it is possible to note how much $l_1$ loss function contributes to the reduction of artifacts affecting the images outputted by the network proposed by Chierchia~et~al.~\cite{chierchia2017sar} trained with our dataset. However, the output generated by SAR-CNN with loss $l_1$ and 17 layers seems to produce strange artifacts in very bright areas, as it can be observed both in the denoised image of Figure~\ref{fig:l1denoised6}.
%Moreover, in our final SAR-CNN for Sentinel-1 image despeckling, the bias that seems to affect areas corresponding to very dark pixels is reduced. The increase in the receptive field of the network let the output be influenced by a larger image region of the input may be the reason behind this improvements. It is worth to point out that details are still very well preserved.

\subsection{Quantitative Comparisons on Images with Simulated~Speckle}
%Simulated noisy SAR images}
\label{sec:er2}
Two common image quality criteria are used to evaluate the quality of despeckling obtained with different methods: the
Peak-signal-to-noise ratio (PSNR), related to the mean squared error,
which~is relevant in terms of evaluation of the estimated
reflectivities (bias and variance of the estimator), and~the
structural similarity (SSIM) which better captures the perceived image
quality. For~each speckle-free image from the testing set, several versions corrupted by synthetic single-look speckle are generated, in~order to report both the PSNR and SSIM mean values, and~their standard
deviations over different noise~realizations.

%
%In image denoising, the quality of the denoised image is usually evaluated by computing Peak-signal-to-noise ratio (PSNR) and structural similarity (SSIM). A bunch of noisy images are generated from the same reference image, and the mean value and the standard deviation of PSNR and SSIM is computed, allowing to evaluate the quality (mean value) and the stability (standard deviation) of denoising.

Seven different images are used in our testing set. We report the
performance of the approaches proposed in this paper as well as
the performance of SAR-BM3D~\cite{parrilli2012nonlocal}, NL-SAR~\cite{deledalle2015nl} and MuLoG+BM3D in Tables~\ref{table:comparison_psnr} and \ref{table:comparison_ssim}.

%In the supporting document, we give the restored images and ratio images for each method.
%
%In our analysis, for each of the 7 test images, a region of interest has been individuated. For each noise-free image, 20 noisy images corrupted by 1-look speckle noise have been generated. The quality of the denoising performance of our pre-trained CNN and our SAR-CNN for Sentinel-1 image despeckling have been compared to that of homomorphic BM3D (with coupled log and exp to adapt the algorithm to handle speckled images), MuLoG+BM3D and homomorphic CNN in terms of PSNR and SSIM computed on the images in amplitude format. The mean value and the variance of these two image quality indexes are listed in Table~\ref{table:comparison_psnr} and Table~\ref{table:comparison_ssim}.

\begin{table}[H]
\centering
\caption{Comparison of denoising quality in terms of PSNR on
amplitude images. For~each ground truth image, 20 noisy
instances are generated. 1$\sigma$ confidence intervals are
given. Per-method averages are given at the bottom.}
\resizebox{\columnwidth}{!}{%
\begin{tabular}{l c c c c c c c}
\toprule
\multirow{2}{*}{\textbf{Images}}      & \multirow{2}{*}{\textbf{Noisy}}            & \multirow{2}{*}{\textbf{SAR-BM3D}}          & \multirow{2}{*}{\textbf{NL-SAR}}            & \multirow{2}{*}{\textbf{MuLoG+BM3D}}       & \multirow{2}{*}{\textbf{MuLoG+CNN}} & \textbf{MuLoG+CNN}  & \multirow{2}{*}{\textbf{SAR-CNN}} \\
&&&&&&\textbf{(Pretrained on SAR)}&\\
\midrule
Marais 1    & 10.05 $\pm$ 0.0141 & 23.56 $\pm$ 0.1335 & 21.71 $\pm$ 0.1258 & 23.46 $\pm$ 0.0794 & 23.39 $\pm$ 0.0608 & 23.63 $\pm$ 0.0678 & \textbf{24.65} $\pm$ 0.0860\\
Limagne     & 10.87 $\pm$ 0.0469 & 21.47 $\pm$ 0.3087 & 20.25 $\pm$ 0.1958 & 21.47 $\pm$ 0.2177 & 21.16 $\pm$ 0.0249 & 21.85 $\pm$ 0.1273& \textbf{22.65} $\pm$ 0.2914\\
Saclay      & 15.57 $\pm$ 0.1342 & 21.49 $\pm$ 0.3679 & 20.40 $\pm$ 0.2696 & 21.67 $\pm$ 0.2445 & 21.88 $\pm$ 0.2195 & 22.77 $\pm$ 0.2403& \textbf{23.47} $\pm$ 0.2276\\
Lely        & 11.45 $\pm$ 0.0048 & 21.66 $\pm$ 0.4452 & 20.54 $\pm$ 0.3303 & 22.25 $\pm$ 0.4365 & 22.17 $\pm$ 0.2702 & 22.97 $\pm$ 0.3671& \textbf{23.79} $\pm$ 0.4908\\
Rambouillet &  8.81 $\pm$ 0.0693 & 23.78 $\pm$ 0.1977 & 22.28 $\pm$ 0.1132 & 23.88 $\pm$ 0.1694 & 23.30 $\pm$ 0.1140 & 23.30 $\pm$ 0.1630& \textbf{24.73} $\pm$ 0.0798\\
Risoul      & 17.59 $\pm$ 0.0361 & 29.98 $\pm$ 0.2638 & 28.69 $\pm$ 0.2011 & 30.99 $\pm$ 0.3760 & 30.85 $\pm$ 0.1844 & 31.03 $\pm$ 0.2008& \textbf{31.69} $\pm$ 0.2830\\
Marais 2    &  9.70 $\pm$ 0.0927 & 20.31 $\pm$ 0.7833 & 20.07 $\pm$ 0.7553 & 21.59 $\pm$ 0.7573 & 21.00 $\pm$ 0.4886 & 22.12 $\pm$ 0.6792& \textbf{23.36} $\pm$ 0.8068\\
\midrule
Average    &  12.00 & 23.17 & 21.99 & 23.62 & 23.39 & 23.95 & \textbf{24.91}\\
\bottomrule
\end{tabular}
}
%The average PSNR and its variance are separated by semicolon.}
\label{table:comparison_psnr}
\end{table}
\unskip
%MDPI: Is the bold necessary? Please confirm and check below.
%ED: we believe it helps to identify the top score

\begin{table}[H]
\centering
\caption{Comparison of denoising quality evaluated in terms of
SSIM on amplitude images. For~each ground truth image, 20
noisy instances are generated. 1$\sigma$ confidence
intervals are given. Per-method~averages are given at the bottom.}
\resizebox{\columnwidth}{!}{%
\begin{tabular}{l c c c c c c c}
\toprule
\multirow{2}{*}{\textbf{Images}}      & \multirow{2}{*}{\textbf{Noisy}}            & \multirow{2}{*}{\textbf{SAR-BM3D}}          & \multirow{2}{*}{\textbf{NL-SAR}}            & \multirow{2}{*}{\textbf{MuLoG+BM3D}}       & \multirow{2}{*}{\textbf{MuLoG+CNN}} & \textbf{MuLoG+CNN}  & \multirow{2}{*}{\textbf{SAR-CNN}} \\
&&&&&&\textbf{(Pretrained on SAR)}&\\
\midrule
Marais 1    &0.3571 $\pm$ 0.0015 &0.8053 $\pm$ 0.0018 &0.7471 $\pm$ 0.0029 &0.8003 $\pm$ 0.0020 &0.7955 $\pm$ 0.0027 &0.8072 $\pm$ 0.0024 & \textbf{0.8333} $\pm$ 0.0016 \\
Limagne     &0.4060 $\pm$ 0.0021 &0.8091 $\pm$ 0.0027 &0.7493 $\pm$ 0.0033 &0.8011 $\pm$ 0.0030 &0.8055 $\pm$ 0.0027 &0.8147 $\pm$ 0.0023& \textbf{0.8327} $\pm$ 0.0029 \\
Saclay      &0.5235 $\pm$ 0.0019 &0.8031 $\pm$ 0.0032 &0.7478 $\pm$ 0.0040 &0.7734 $\pm$ 0.0034 &0.7956 $\pm$ 0.0033 &0.8156 $\pm$ 0.0030& \textbf{0.8314} $\pm$ 0.0024 \\
Lely        &0.3654 $\pm$ 0.0013 &0.8473 $\pm$ 0.0023 &0.8062 $\pm$ 0.0023 &0.8552 $\pm$ 0.0025 &0.8659 $\pm$ 0.0019 &0.8703 $\pm$ 0.0018& \textbf{0.8856} $\pm$ 0.0019 \\
Rambouillet &0.2886 $\pm$ 0.0017 &0.7831 $\pm$ 0.0028 &0.7364 $\pm$ 0.0031 &0.7798 $\pm$ 0.0029 &0.7706 $\pm$ 0.0095 &0.7821 $\pm$ 0.0073& \textbf{0.8002} $\pm$ 0.0026 \\
Risoul      &0.4362 $\pm$ 0.0017 &0.8306 $\pm$ 0.0024 &0.7671 $\pm$ 0.0028 &0.8345 $\pm$ 0.0030 &0.8291 $\pm$ 0.0027 &0.8341 $\pm$ 0.0024& \textbf{0.8493} $\pm$ 0.0018 \\
Marais 2    &0.2628 $\pm$ 0.0017 &0.8506 $\pm$ 0.0026 &0.8222 $\pm$ 0.0022 &0.8561 $\pm$ 0.0025 &0.8594 $\pm$ 0.0111 &0.8677 $\pm$ 0.0097& \textbf{0.8866} $\pm$ 0.0025 \\
\midrule
Average    &  0.3771 & 0.8184 & 0.7680 & 0.8143 & 0.8173 & 0.8273 & \textbf{0.8460}\\
\bottomrule
\end{tabular}
}

%The average SSIM and its variance are separated by semicolon.}
\label{table:comparison_ssim}
\end{table}

\begin{figure}[H]
\centering
\includegraphics[width=15.4 cm]{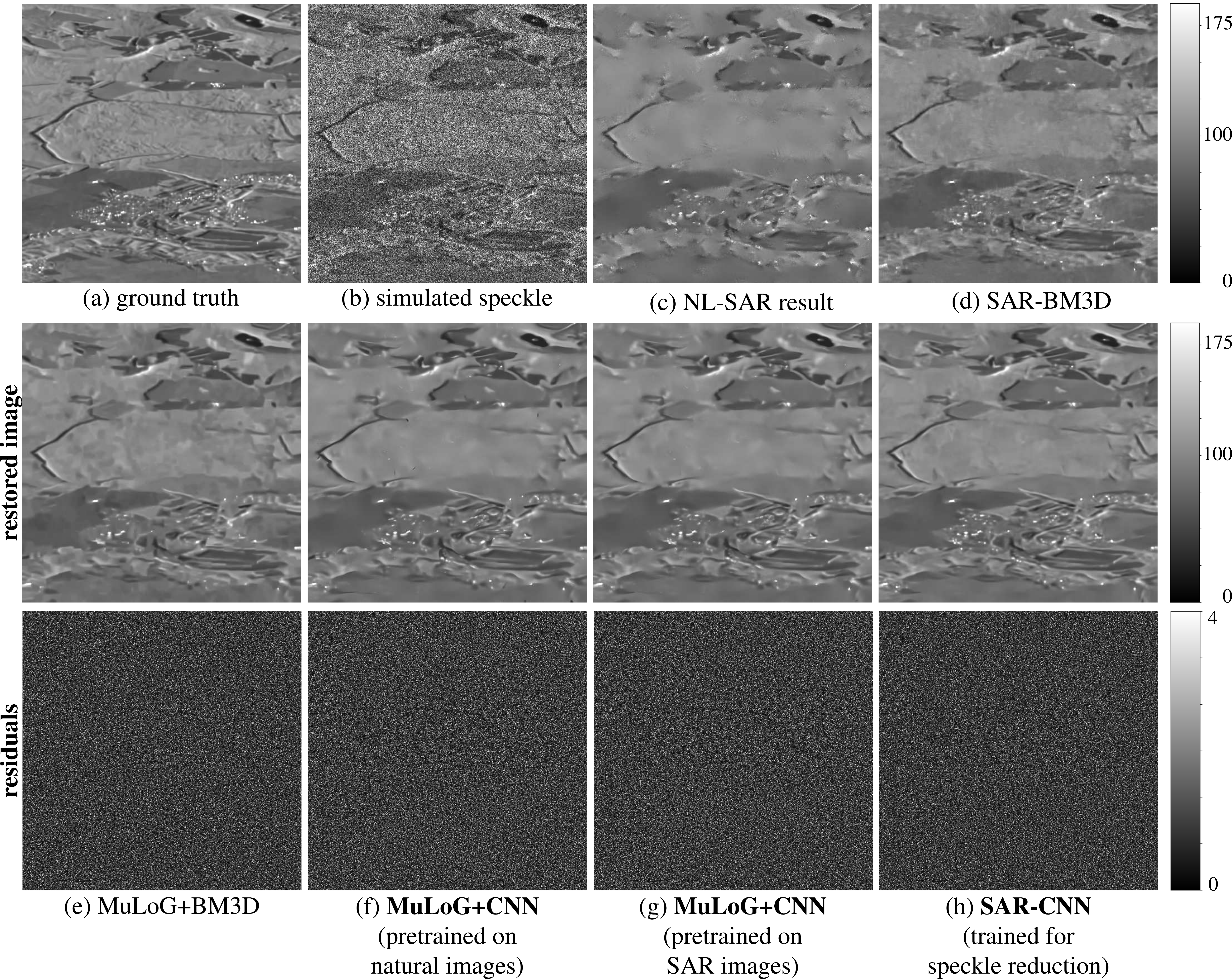}
\caption{Restoration results and ratio images (noisy/denoised) on an image from the testing set. %By repeating the restoration on 20 independent speckle simulations, the~bias and relative standard deviation are computed and displayed, at~each pixel, in~the last two rows.
}
%presentation of the bias and relative standard deviation, computed from the restoration of 20 independent speckle simulations, and a ratio image (noisy/denoised).}
\label{fig:comparisonmarais1}
\end{figure}

To qualitatively evaluate the quality of denoising, we display in
Figure~\ref{fig:comparisonmarais1} the results obtained by the
different methods on image ``Marais 1''. To~better analyze the results, %beyond the restored images,
the residual intensity images (i.e.,~the ratio between the noisy and the restored images) are displayed below the restored images. Almost no structure can be identified by visual analysis of the residual images, which~means that the compared methods preserve very well the geometrical content of the original images (limited~over-smoothing).

\vspace{-6pt}

It can be observed both on the quantitative results reported in the
tables and in the qualitative analysis that the CNN methods (the two versions of MuLoG+CNN and SAR-CNN) perform better than MuLoG+BM3D which we use as reference algorithm in SAR despeckling before the introduction of CNN
methods.
SAR-CNN removes speckle from the images while preserving the
details, such~as edges, at~the cost of introducing small but
noticeable artifacts in homogeneous areas. Instead, MuLoG+BM3D and MuLoG+CNN generate blurry edges, over-smoothing some areas where
the details are lost, even~when the CNN is pre-trained on SAR images. This can be observed comparing the denoised
images of Figure~\ref{fig:comparisonmarais1}, where SAR-CNN
preserves better the details of the urban area at the bottom of the
image compared to the three other denoising methods. The~quality of the
details can be attributed to the richness of information captured by
the network when learning on many SAR image~patches.

\subsection{Despeckling of Real Single-Look SAR Images: How to Handle~Correlations} \label{sec:er3}
In this section, the~denoising performance of SAR-CNN for Sentinel-1 image despeckling and MuLoG+CNN are evaluated on real single-look SAR images acquired during Sentinel-1 mission. To~test our deep learning-based denoiser, we focused on some of the areas of the images analyzed in the above tables, picking one of the multitemporal instances used to generate the ground truth images for the training of SAR-CNN and making sure that these areas do not belong to the training~set.

Unlike synthetically generated noisy SAR images, in~real acquisitions,
pixels are spatially correlated. SAR images undergo an apodization
(and over-sampling) process
~\cite{pastina2007effect,stankwitz1995nonlinear,abergel2018subpixellic}
aimed at reducing the sidelobes of strong targets, by~introducing some
spectral weighting. %Like most of the statistical methods for speckle reduction~\cite{argenti2013tutorial}, our SAR-CNN assumes that the speckle-noise is an uncorrelated process affecting noise-free data. However, this hypothesis does not hold in practice due to the frequency response of SAR systems, impacting the denoising performance. Indeed, if~correlation is not taken into account and a method is applied directly on correlated data, noticeable artifacts may be introduced in the denoised image. When dealing with real SAR acquisition, we propose to reduce the effect of the correlation by subsampling the image at hand, resulting in a loss of resolution. A~subsampling factor of 2 is a good trade-off between the reduction of correlation ensuring good denoising quality and the loss of correlation. The~despeckled images are then over-sampled through bilinear interpolation to reconstruct an image at the original pixel resolution.
Thus, we subsample real SAR acquisitions by a factor of 2, as~proposed in~\cite{dalsasso2020handle}. As~already observed in the case of synthetic SAR data, the~images that are restored with the proposed methods show significant improvements on the denoising performance over the reference despeckling algorithm MuLoG+BM3D, with~SAR-CNN being the one that provides the best visual result. Even when compared to SAR-BM3D and NL-SAR, which do not require the image to undergo a downsampling step, our results are more good-looking, with~a better preservation of fine structures. NL-SAR, indeed, gives its best in polarimetric and interferometric~configurations.

\begin{figure}[H]
\centering
\includegraphics[width=15.5 cm]{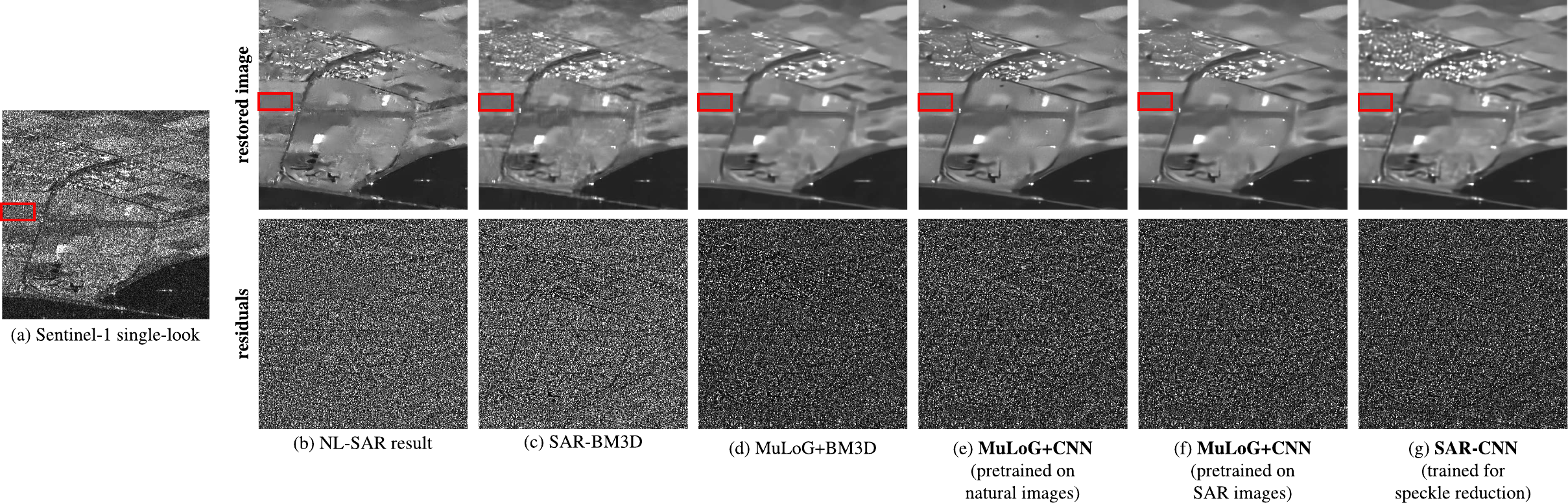}
\caption{Despeckling results on a single-look Sentinel-1 image: (\textbf{a})~the single-look SAR image, (\textbf{b})~a~restoration result obtained by NL-SAR, (\textbf{c}) restoration obtained by SAR-BM3D, (\textbf{d}) restoration obtained by MuLoG+BM3D,
(\textbf{e}) restoration obtained with the pre-trained CNN and MuLoG framework,
(\textbf{f}) restoration obtained with a CNN trained on SAR images and MuLoG framework,
(\textbf{g}) restoration obtained with our network trained for speckle removal on Sentinel-1 images. Images at the bottom row give the residuals of the restoration results.
The red box indicates the area that has been used to estimate the equivalent number of looks to evaluate the quality of denoising.}
\label{fig:reallely}
\end{figure}

Figure~\ref{fig:reallely} shows the restoration results and the residual images (obtained by forming the ratio noisy/denoised) obtained with the different methods. In~contrast to the simulated case, some~structures can be visually identified in the residual images. These structures correspond to thin roads. The~downsampling operation necessary to remove the speckle correlations make the preservation of these structures very difficult for all methods. SAR-CNN seems to be the most effective at preserving those structures. Indeed, the~residual image does not present clear textures nor constant areas, meaning that no structures have been removed and that noise there is no noise left in the restored image (the~residual image represents what is suppressed from the noisy image). Visual analysis of the restored images compared to the real image also seems to indicate fewer artifacts with~SAR-CNN.
%MDPI: There is no (h) and (i) in Figure 8.
%ED: the caption has been modified

Since the ground-truth reflectivity is not available, to~measure the performance of the proposed method the Equivalent Number of Looks is estimated on manually selected homogeneous areas. The~homogeneous regions chosen for the ENL estimation are shown with red boxes, and~the ENL values are given in Table~\ref{table:sentinel1enl}.
\begin{table}[H]
\centering
\caption{ENL estimation on denoised 1-look Sentinel-1 SAR acquisitions and on a TerraSAR-X~image.}
\resizebox{\columnwidth}{!}{%
\begin{tabular}{l c c c c c c}
\toprule
\multirow{2}{*}{\textbf{Images}} & \multirow{2}{*}{\textbf{SAR-BM3D}}          & \multirow{2}{*}{\textbf{NL-SAR}}            & \multirow{2}{*}{\textbf{MuLoG+BM3D}}       & \multirow{2}{*}{\textbf{MuLoG+CNN}} & \textbf{MuLoG+CNN}  & \multirow{2}{*}{\textbf{SAR-CNN}} \\
&&&&&\textbf{(Pretrained on SAR)}&\\
\midrule
\bf Sentinel-1:\\
%\midrule
Marais 1    & 226.48 & 165.24 & 132.30 & \textbf{288.70} & 210.17 & 177.72\\
Lely        & 166.60 & 75.19 & \textbf{349.32} & 82.24 & 145.07 & 289.03\\
Rambouillet & 262.47 & 171.42 & 139.62 & \textbf{413.09} & 383.81  & 295.30\\
Marais 2    & 119.99 & \textbf{213.45} & 84.67 & 146.33 & 182.44 & 206.93\\
\midrule
\bf TerraSAR-X:\\
%\midrule
Saint Gervais    & 40.01 & 39.70 & 39.37 &45.18 & \textbf{129.66}& 59.21\\
\bottomrule
\end{tabular}
}\label{table:sentinel1enl}
\end{table}

\begin{figure}[H]
\includegraphics[width=15.5 cm]{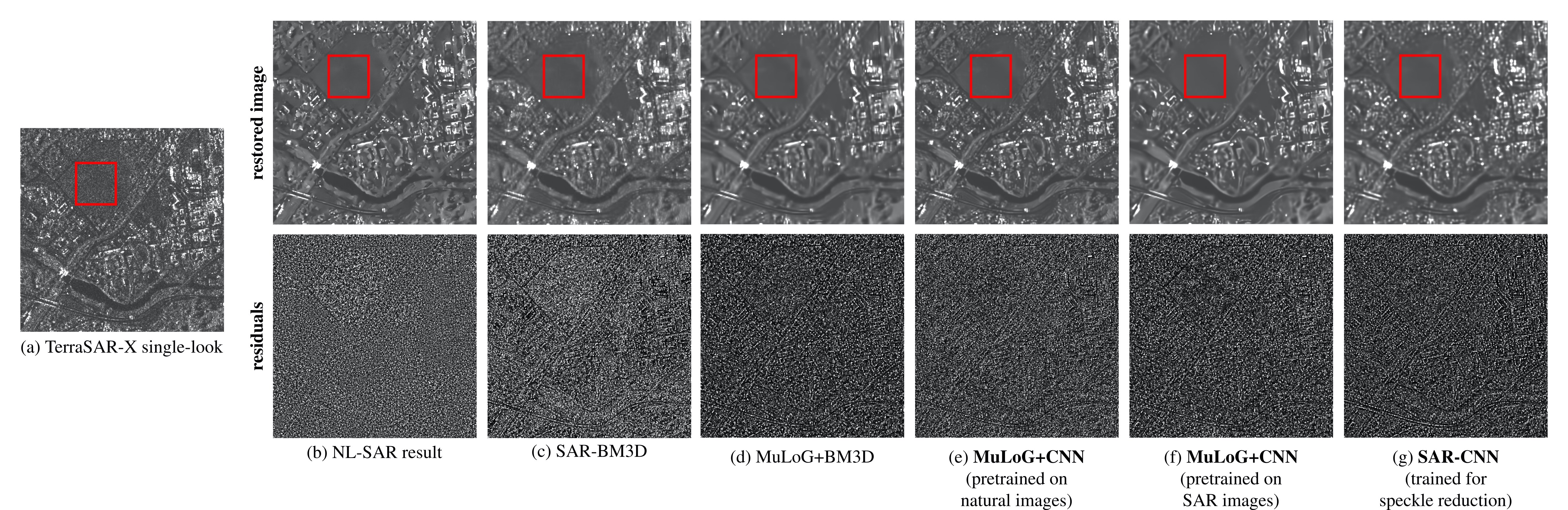}
\caption{Despeckling results on a single-look TerraSAR-X image: (\textbf{a}) a TerraSAR-X image in Stripmap mode (1 m $\times$ 2 m spatial resolution), (\textbf{b}) a restoration result obtained by NL-SAR, (\textbf{c}) restoration obtained by SAR-BM3D, (\textbf{d}) restoration obtained by MuLoG+BM3D,
(\textbf{e}) restoration obtained with the pre-trained CNN and MuLoG framework,
(\textbf{f}) restoration obtained with a CNN trained on SAR images and MuLoG framework,
(\textbf{g}) restoration obtained with our network trained for speckle removal on Sentinel-1 data. Images at the bottom row give the residuals of the restoration results.
As in Figure~\ref{fig:reallely}, the~red box indicates the area used to estimate the equivalent number of looks.}
\label{fig:realsgervais}
\end{figure}

Then, this analysis has been extended to a TerraSAR-X acquisition. In~this case, the~study aims at assessing the generalization capabilities of the
trained SAR-CNN on images from a different sensor and a different
spatial resolution. While MuLoG+CNN is not dedicated to a specific sensor (see Table~\ref{table:comparison_runtime}), SAR-CNN is sensor specific. As~such, to~give its best on TerraSAR-X images, it should be retrained using images acquired from TerraSAR-X: a dataset can be constructed as described in Section~\ref{sec:pm31}. As~it can be seen from Figure~\ref{fig:realsgervais}, MuLoG+CNN seems
the approach that provides visually the best results. The~estimated
ENL indicate that all despeckling methods are very effective in
homogeneous regions. It seems that MuLoG+CNN produces
an image with a slightly better perceived resolution, which
may indicate a better generalization property compared to
SAR-CNN. While the latter would benefit from a dedicated training on TerraSAR-X images, it achieves satisfying performance on this data without any fine-tuning, which indicates that it could be readily applied to images from other sensors in the absence of time series to build a ground truth for retraining. The~analysis of the residual images, like in Figure~\ref{fig:reallely}, indicates that thin linear structures are attenuated in restored images due to the downsampling~step.

\unskip

\begin{table}[H]
\centering
\caption{Time for despeckling a $500\times500$ clip. Experiments were carried out with an Intel Xeon CPU at 3.40 GHz and an Nvidia K80 GPU. For~NL-SAR, the~radius of the smallest/largest search window size is set to 1/20 and the half-width of the smallest/largest patches as 0/10.}
\begin{tabular}{l c c c c}
\toprule
\textbf{SAR-BM3D}           & \textbf{NL-SAR}            & \textbf{MuLoG+BM3D}       & \textbf{MuLoG+CNN} & \textbf{SAR-CNN} \\
\midrule
73.89 s& 116.28 s & 59.82 s& 80.43 s& \textbf{0.19 s} \\
\bottomrule
\end{tabular}
%The average PSNR and its variance are separated by semicolon.}
\label{table:comparison_runtime}
\end{table} %MDPI:We insert it close to where it is first mentioned in the text.

\section{Discussion} \label{sec:cl}
%In this paper, we investigated the use of state-of-the-art CNNs in suppressing speckle from SAR images. Two approaches have been proposed. In the first one, a pre-trained CNN for Gaussian denoising is used inside the MuLoG scheme. In the second one, a new learning strategy for the generation of reliable ground-truth images has been deployed to train a CNN. These two methods are tested on simulated SAR images. Our two approaches have shown to reach state-of-the-art results, outperforming popular algorithms in the field of speckle removal. Promising results have been also obtained on real images, where a downsampling step is required before applying denoising to reduce the effect of spatial correlation. Comparing these two methods using CNN, training an \textit{ad-hoc} CNN model achieves better performance. By sacrificing a little performance, the method using the pre-trained model can handle the SAR images in different numbers of looks and different polarization.

%[LOIC: the conclusion could be improved in particular with respect to
%the new title I suggested. It seems to me that the
%resolution of the real Sentinel-1 and TerraSAR-X images is better
%preserved with MuLoG+CNN, am I wrong?]

A CNN trained to suppress additive white Gaussian noise encodes a very
generic model of natural images in the form of a proximal operator
related to an implicit prior. Like other models of natural images that
were successfully applied to the problem of speckle reduction in SAR
imaging (wavelets, total variation), they are relevant to SAR imagery
because they capture structures (points,~edges, corners) and textures.
Yet, the~specificities of SAR images make it beneficial to train a
model specifically on SAR images. This is done naturally by
patch-based methods that use the content of the image itself as a model
(repeating patches).
In this paper, we have shown a considerable improvement with a CNN model
trained on SAR images provided that a high-quality training set is
built, enough layers are used to capture large scale structures and an
adequate loss function is selected.
Moreover, once trained, SAR-CNN exhibits the best runtime performance when using a GPU (see~Table~\ref{table:comparison_runtime}).
When considering the real-life case of partially correlated speckle
or images from different sensors, the~plugging of a network trained on natural images
in a SAR adapted framework like MuLoG presents better generalization~properties.

The extension to multi-channel SAR images represents a real challenge.
Speckle reduction in multi-channel images requires modeling the
correlations between channels (the interferometric and polarimetric
information). In~order to \emph{learn} those correlations directly
from the data, a~dataset that contains \emph{speckle-free} images covering the whole diversity of polarimetric
responses, interferometric phase differences and the whole range of
coherences for typical geometrical structures (points, lines,
corners, homogeneous regions, textured regions) must be
formed. Needless to say, this is far more challenging than collecting
single-channel SAR images to cover only the diversity of geometric
structures. Failure to correctly include all cases in the training set
implies that the network, instead of performing a high-dimensional
interpolation, performs a high-dimensional extrapolation, which puts
the user at high risk of experiencing large prediction~errors.

This difficulty justifies the relevance of using pre-trained networks
within MuLoG framework which has been designed to apply single-channel
restoration methods to multi-channel SAR images.
%
%Indeed, the increase of the data dimensionality requires a much larger
%training set in order to faithfully cover the diversity of images met in
%practice. The application of a pre-trained network, used as a rich model
%of the structures and textures found in natural images, is then an
%alternative to training a dedicated network that is much easier to apply
%by using the multi-channel framework MuLoG.
A~summary of the advantages and drawbacks of the proposed methods are
reported in Table~\ref{table:prosandcons}. Only~a~little gain in performance is observed when, considering MuLoG+CNN, the~network is pre-trained on SAR images. Thus, if~creating a dataset is possible it is advisable to use it to train an end-to-end model such as~SAR-CNN.

To offer the possibility to use the presented SAR-CNN for testing and comparison, we release an open-source code of the network trained on our dataset: {\url{https://gitlab.telecom-paris.fr/RING/SAR-CNN}}.
%REMOTE SENSING do not have footnote.
Indeed, replicating results of a published work is not an easy task and may represents up to months of work. Therefore, by~sharing our code, we hope to help other researchers and users of SAR images to easily apply our CNN-based denoiser on single-look Sentinel-1 images, and~possibly compare the restoration performance with their own methods.
%Moreover, training a model would require to build a dataset of reference images.

\begin{table}[H]
\centering
\caption{Advantages and disadvantages of the CNN-based despeckling strategies considered in this~paper.}
\begin{tabular}{lcc}
\toprule
\multirow{3}{*}{MuLoG+CNN} & Pros &
\begin{minipage}{4in}
\vskip 1pt
\begin{itemize}[leftmargin=*,labelsep=5.8mm]
\item No specific training is needed\\ (improvement is small when the CNN is trained on SAR~images)
\item Straightforward adaptation to multiple looks
\item Straightforward generalization to polarimetric and/or interferometric SAR images
\item Straightforward adaptation to different SAR sensors
\end{itemize}
\vskip 1pt
\end{minipage}  \\ \cmidrule{2-3}
& Cons &
\begin{minipage}{4in}
\vskip 1pt
\begin{itemize}[leftmargin=*,labelsep=5.8mm]
\item High runtime due to its iterative procedure
\end{itemize}
\vskip 1pt
\end{minipage} \\
\midrule
\multirow{6}{*}{SAR-CNN}     & Pros &
\begin{minipage}{4in}
\vskip 1pt
\begin{itemize}[leftmargin=*,labelsep=5.8mm]
\item Provides the highest performances
\item Fastest runtime performances once the network is trained
\item Possible adaptation to multiple looks (requires~re-training)
\item Possible adaptation to different SAR sensors (requires~re-training)
\end{itemize}
\vskip 1pt
\end{minipage} \\ \cmidrule{2-3}
& Cons &
\begin{minipage}{4in}
\vskip 1pt
\begin{itemize}[leftmargin=*,labelsep=5.8mm]
\item Requires a training: formation of a dataset of speckle-free SAR images
\item %Application to differen
Generalization to multi-channel SAR images (polarimetric and/or interferometric) raises dimensionality issues: very large training set to sample the diversity of radar images
%other applications subject to the possibility of training the network with a proper dataset
%				\item Increase of dimensionality when dealing with POLinSAR data makes it difficult to train a CNN model
\end{itemize}
\vskip 4pt
\end{minipage}  \\
\bottomrule
\end{tabular}
%The average PSNR and its variance are separated by semicolon.}
\label{table:prosandcons}
\end{table}

\section{Conclusions} \label{sec:conclusions}
With the new generation of sensors orbiting around Earth, access to long time-series of SAR images is improving. Given the increasing interest towards the use of deep learning algorithms in SAR despeckling, in~this paper it is described a procedure to generate ground truth images that can be applied in a systematic way to produce large training sets formed by pairs of high-quality speckle-free images and simulated speckled~images.

In a future work, it would be interesting to train the networks on actual single-look SAR images in order to account for spatial correlations of the speckle. This, however, would require a method to produce a high quality ground-truth image for each single-look observation. Restoration methods that exploit long time-series of SAR images~\cite{zhao2019ratio} may pave the way to producing such training~sets.
\vspace{6pt}

\authorcontributions{Conceptualization: F.T and L.D.; methodology and formal analysis: E.D., X.Y., L.D., F.T; writing: E.D., X.Y., L.D., F.T.; review and editing: E.D., L.D., F.T.; supervision: L.D., F.T., W.Y., software: L.D., E.D. All authors have read and agreed to the published version of the manuscript.}
%MDPI: For research articles with several authors, a~short paragraph specifying their individual contributions must be provided. The~following statements should be used ``conceptualization, X.X. and Y.Y.; methodology, X.X.; software, X.X.; validation, X.X., Y.Y. and Z.Z.; formal analysis, X.X.; investigation, X.X.; resources, X.X.; data curation, X.X.; writing--original draft preparation, X.X.; writing--review and editing, X.X.; visualization, X.X.; supervision, X.X.; project administration, X.X.; funding acquisition, Y.Y.'', please turn to the  \href{http://img.mdpi.org/data/contributor-role-instruction.pdf}{CRediT~taxonomy} for the term explanation. Authorship must be limited to those who have contributed substantially to the work reported.

%%%%%%%%%%%%%%%%%%%%%%%%%%%%%%%%%%%%%%%%%%
\funding{This research received no external funding.}
%MDPI: Please add: ``This research received no external funding'' or ``This research was funded by NAME OF FUNDER grant number XXX.'' and  and ``The APC was funded by XXX''. Check carefully that the details given are accurate and use the standard spelling of funding agency names at \url{https://search.crossref.org/funding}, any errors may affect your future~funding.

%%%%%%%%%%%%%%%%%%%%%%%%%%%%%%%%%%%%%%%%%%
%\acknowledgments{\hl{Text}.}
%MDPI: In this section you can acknowledge any support given which is not covered by the author contribution or funding sections. This may include administrative and technical support, or~donations in kind (e.g., materials used for experiments).

%%%%%%%%%%%%%%%%%%%%%%%%%%%%%%%%%%%%%%%%%%
\conflictsofinterest{The authors declare no conflict of interest.}
\reftitle{References}

%\begin{thebibliography}{999}
% Reference 1
%\bibitem[Author1(year)]{ref-journal}
%Author1, T. The title of the cited article. {\em Journal Abbreviation} {\bf 2008}, {\em 10}, 142--149.
% Reference 2
%\bibitem[Author2(year)]{ref-book}
%Author2, L. The title of the cited contribution. In {\em The Book Title}; Editor1, F., Editor2, A., Eds.; Publishing House: City, Country, 2007; pp. 32--58.
%\bibliography{ref}
%\end{thebibliography}

% The following MDPI journals use author-date citation: Arts, Econometrics, Economies, Genealogy, Humanities, IJFS, JRFM, Laws, Religions, Risks, Social Sciences. For those journals, please follow the formatting guidelines on http://www.mdpi.com/authors/references
% To cite two works by the same author: \citeauthor{ref-journal-1a} (\citeyear{ref-journal-1a}, \citeyear{ref-journal-1b}). This produces: Whittaker (1967, 1975)
% To cite two works by the same author with specific pages: \citeauthor{ref-journal-3a} (\citeyear{ref-journal-3a}, p. 328; \citeyear{ref-journal-3b}, p.475). This produces: Wong (1999, p. 328; 2000, p. 475)

%=====================================
% References, variant B: external bibliography
%=====================================
%\externalbibliography{yes}
%\bibliography{your_external_BibTeX_file}

%%%%%%%%%%%%%%%%%%%%%%%%%%%%%%%%%%%%%%%%%%
%% optional
%\sampleavailability{Samples of the compounds ...... are available from the authors.}

%% for journal Sci
%\reviewreports{\\
%Reviewer 1 comments and authors’ response\\
%Reviewer 2 comments and authors’ response\\
%Reviewer 3 comments and authors’ response
%}

%%%%%%%%%%%%%%%%%%%%%%%%%%%%%%%%%%%%%%%%%%
\end{document}